\pdfoutput=1
\documentclass[11pt]{article}

\usepackage[]{EMNLP2023}

\usepackage{times}
\usepackage{latexsym}

\usepackage[T1]{fontenc}
\usepackage[utf8]{inputenc}

\usepackage{microtype}

\usepackage{inconsolata}

\usepackage{multirow}
\usepackage{booktabs}
\usepackage{caption}
\usepackage{subcaption}
\usepackage{xspace}
\usepackage{amsmath}
\usepackage{amssymb}
\usepackage{newtxmath}
\usepackage{bbm}
\usepackage{graphicx}
\usepackage{tabularx}
\usepackage{bbding}
\usepackage{pifont}
\usepackage{url}
\usepackage[ruled,vlined]{algorithm2e}
\usepackage{diagbox}
\usepackage{makecell}
\usepackage{colortbl}
\usepackage{bbm}
\usepackage{color}
\usepackage{xcolor}
\usepackage{hhline}
\usepackage{soul}

\newcommand{\paratitle}[1]{\vspace{1.5ex}\noindent\textbf{#1}}
\newcommand{\ie}{\emph{i.e.,}\xspace}

\newcommand{\eg}{\emph{e.g.,}\xspace}

\newcommand{\ignore}[1]{}

\title{Not Everything is All You Need: Toward Low-Redundant Optimization for \\ Large Language Model Alignment}
\author{
    \textbf{Zhipeng Chen\textsuperscript{{1}}\thanks{\llap{}\:\:\:Equal contribution. },~
	        Kun Zhou\textsuperscript{{2}}\footnotemark[1],~
	        Wayne Xin Zhao\textsuperscript{{1}}\thanks{\llap{}\:\:\:Corresponding author. }}, \\
                \textbf{Jingyuan Wang\textsuperscript{{3}}
                \and Ji-Rong Wen\textsuperscript{{1},{2}}}\\
	\textsuperscript{1}Gaoling School of Artificial Intelligence, Renmin University of China.\\
	\textsuperscript{2}School of Information, Renmin University of China.\\
        \textsuperscript{3}School of Computer Science and Engineering, Beihang University. \\
        \texttt{zhipeng\_chen@ruc.edu.cn,francis\_kun\_zhou@163.com,batmanfly@gmail.com}
}

\begin{document}
\maketitle

\begin{abstract}
Large language models (LLMs) are still struggling in aligning with human preference in complex tasks and scenarios.
They are prone to overfit into the unexpected patterns or superficial styles in the training data.
%researchers have proposed several methods to align LLMs to human preferences.
%However, according to the existing work, 
We conduct an empirical study that only selects the top-10\% most updated parameters in LLMs for alignment training, and see improvements in the convergence process and final performance.
It indicates the existence of redundant neurons in LLMs for alignment training.
%LLMs might overfit to the unexpected patterns of training data, because of the redundant updates in the alignment process.
%To verify this perspective, we have conducted an empirical study and observed several redundant elements in the training process, \eg redundant neurons and tokens, which reduce the training efficiency, including convergence speed and downstream task performance.
To reduce its influence, we propose a low-redundant alignment method named \textbf{ALLO}, focusing on optimizing the most related neurons with the most useful supervised signals.
Concretely, we first identify the neurons that are related to the human preference data by a gradient-based strategy, then identify the alignment-related key tokens by reward models for computing loss.
%focusing on eliminating the redundancy in the alignment process via leveraging a gradient-based method to select the neurons related to the downstream scenario and the token-level rewards to identify the important tokens in the training data.
Besides, we also decompose the alignment process into the forgetting and learning stages, where we first forget the tokens with unaligned knowledge and then learn aligned knowledge, by updating different ratios of neurons, respectively.
Experimental results on 10 datasets
%complex reasoning tasks and instruction following tasks 
have shown the effectiveness of ALLO.
% Our code and data will be publicly released.
Our code and data are available at \url{https://github.com/RUCAIBox/ALLO}.

\end{abstract}

\section{Introduction}

%Recently, owing to the scaling law, large language models (LLMs) (\eg GPT-4~\cite{GPT-4}, Claude 3~\cite{claude3}, LLaMA 3~\cite{llama3}) can exhibit superior capabilities on various tasks~\cite{LLMsurvey}.
%For their broad application in the real world, 
Alignment with human preferences has become a desired property of LLMs~\cite{anthropic_3h,instructgpt,LLMsurvey}, \eg helpfulness, honesty, and harmlessness, 
and reinforcement learning from human feedback (RLHF)~\cite{Christiano-NeurIPS-2017-Deep,moss_rlhf} is a widely utilized technique for achieving it.
%improving human alignment of LLMs.
Typically, RLHF aims to fine-tune LLMs on human preference data, to maximize and minimize the likelihood of generating the positive and negative responses, respectively.
%Efforts to fine-tune LLMs  have been widely studied in recent work, including supervised fine-tuning (SFT)~\cite{instructgpt,flant5} and , LLMs can further enhance their abilities and adapt to specific domains.
After RLHF training on corresponding datasets, LLMs can better follow user instructions~\cite{instructgpt}, solve complex problems~\cite{math-shepherd}, and generate unbiased responses~\cite{Bai-2022-Training-arxiv}.
% (\eg solving complex problems or providing specific information)~\cite{} and align with human preferences (\eg generating unbiased content or reducing hallucination)~\cite{}.

\begin{figure}[t]
    \centering
    \includegraphics[width=0.48\textwidth]{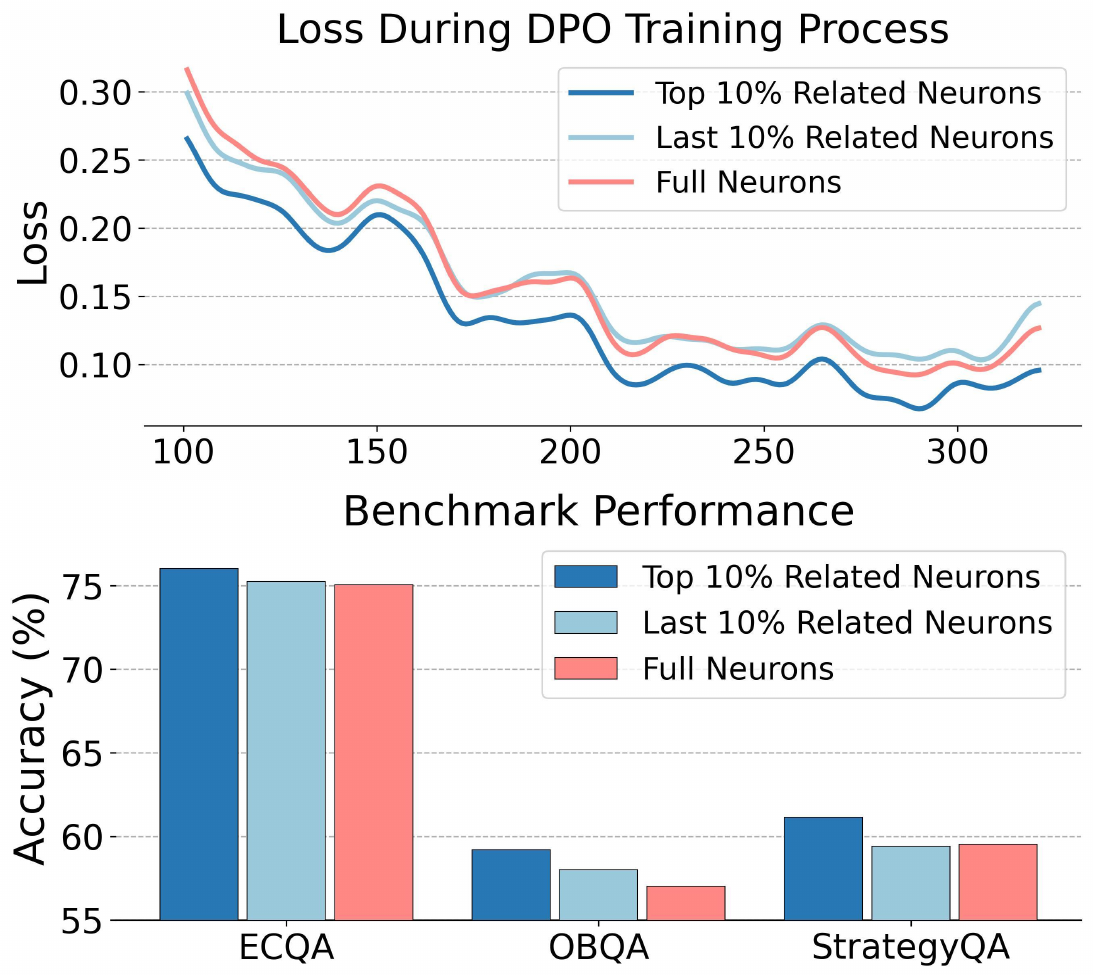}
    \caption{Training loss curve and benchmark performance of QA tasks using different trainable neurons in LLM. We perform alignment training using DPO~\cite{dpo} on ECQA and QASC. The top/last-10\% related neurons are selected based on the accumulated gradients during DPO training.}
    % has been adopted as the training method in this empirical study.}
    \label{intro}
\end{figure}

However, it is hard to train a well-aligned LLM for complex tasks and scenarios~\cite{Feng-2024-Towards-arxiv,Gekhman-2024-Does-arxiv}.
The key issue is that LLMs might overfit into the unexpected patterns or superficial styles in the human preference data (\ie pairs of positive-negative response)~\cite{Du-2024-Shortcut-ACM}.
It is the side effect of their powerful learning capability derived from the large-scale trainable parameters~\cite{Song-2024-Large-arxiv,Meng-2024-ALPS-arxiv}.
Recently, a surge of work~\cite{LotteryTicketHypothesis,Wang-2024-Sharing-arxiv} has found that each neuron (\ie one of the trainable values of the parameter matrixes in LLMs) is relevant with special knowledge, and the neurons in LLMs are generally sparsely activated.
Inspired by it, we consider whether the irrelevant neurons exist in LLMs during the alignment process and the full-parameter trained LLMs might lead to \textbf{\emph{redundant updates}} on alignment-irrelevant neurons.
Thus, we conduct the empirical experiment using DPO algorithm, where we only update the top/last-10\% neurons according to their accumulated gradient values, which indicate the relevance between the current neuron and the downstream scenario~\cite{Pruthi-2023-Estimating-neurips}.
As shown in Figure~\ref{intro}, with the top-10\% trainable neurons, LLMs can converge faster and achieve better performance than optimizing all the neurons.
It indicates the existence of alignment-irrelevant neurons and redundant updates in DPO training, affecting the convergence and final performance.
Besides the redundant neurons, previous work~\cite{Lin-2024-Rho-arxiv} has shown the existence of redundant tokens in training data.

%in the recent study~\cite{}, the stability of the LLM training process has become a critical issue, especially in the alignment process (\eg PPO, DPO).
%One of the possible reasons is that there are several redundant elements (\eg neurons in LLMs, training data) in the LLM training process, which might guide LLMs to learn something trivial and influence the training efficiency (\eg convergence speed or performance on downstream tasks)~\cite{}.

To reduce the influence of redundant updates, we aim to prune the alignment training, to focus on optimizing the most related neurons with the most useful supervised signals.
Concretely, we first identify the neurons that are related to the human preference data, based on the accumulation values of gradients.
Second, we identify the key tokens about human preference, and only compute loss on them for optimizing the alignment-related neurons.
%via a token-level reward model, which is trained by distilling GPT-4.
%During training, we only compute loss on the key tokens, and optimize the 
In this way, we perform a \textbf{\emph{low-redundant optimization}} for aligning LLMs with humans, which reduces the redundancy of learning irrelevant tokens and training irrelevant neurons, avoiding their effect on alignment performance.

However, the involved tokens and neurons are not always consistent in the alignment objectives, since the alignment training focuses on both removing the unaligned knowledge and learning the aligned one.
Therefore, we decompose the alignment process into the \textbf{\emph{forgetting and learning stages}}, and adapt the low-redundant optimization strategy on them.
For the forgetting stage, relatively fewer neurons are trained by unlearning algorithm~\cite{npo} to forget the unaligned knowledge, and we leverage a token-level reward model~\cite{rlmec} to identify the unaligned tokens required to be focused.
For the learning stage, we train more neurons using the DPO algorithm, and also utilize its reward score to select the key tokens.

%To address these problems, we focus on training the neurons which are most relative to the knowledge of downstream tasks, and performing fine-grained supervision during the training process in LLM alignment.
In this work, we proposed an \textbf{AL}ignment method with \textbf{L}ow-Redundant \textbf{O}ptimization~(ALLO) to fine-tune LLMs.
In ALLO, we first identify the most important neurons by training a reference model. 
Then, for the forgetting and learning stages, we utilize NPO and DPO algorithms with unlearning and learning losses on corresponding key tokens respectively, to optimize the identified important neurons.
To comprehensively assess the effectiveness of ALLO, we conduct extensive experiments on three downstream scenarios, \ie question answering, mathematical reasoning, and instruction following, totally 10 datasets.
Experiment results show that ALLO mostly outperforms competitive human alignment methods (\eg SFT~\cite{instructgpt}, DPO~\cite{dpo}, PPO~\cite{ppo}), achieving a 9.7\% maximum relative improvement over vanilla DPO.
%which verifies ALMEC can enhance the training efficiency in LLM alignment process.

\section{Related Work}
We introduce the related work from the perspectives of large language models, LLMs alignment, and unlearning of LLMs.

\paratitle{Large Language Models.}
LLMs have shown remarkable performance on various tasks~\cite{qwen2,llama3,phi2}.
Generally, the training process of LLMs includes three stages, \ie pre-training, supervised fine-tuning (SFT), and alignment~\cite{instructgpt,llama2}.
In the training process, previous work has selected valuable data to train the LLMs via leveraging gradient~\cite{Xia-2024-LESS-arxiv} or perplexity~\cite{Lin-2024-Rho-arxiv,Xie-2023-DoReMi-neurips}, 
Besides, synthetic training data from powerful LLMs (\eg GPT-4, Claude 3) has been widely utilized for improving the weak LLMs~\cite{Xu-2023-WizardLM-arxiv,allal-2024-cosmopedia-hf,Liu-2024-Best-arxiv}, especially for specific scenarios (\eg mathematical tasks or code synthesis tasks)~\cite{Yue-2023-MAmmoTH-arxiv,zhou-2024-jiuzhang3-arxiv}.
However, given the large expenses of the LLM training, existing work~\cite{lora,prefix-tuning,qlora} has revealed that training only a small number of the parameters can achieve comparable performance with whole-parameters training.
In this work, we focus on the alignment stage and leverage the low-redundant optimization to improve the existing LLMs.

\paratitle{LLMs Alignment.}
RLHF is a critical algorithm of LLM alignment~\cite{Christiano-NeurIPS-2017-Deep}, usually leveraged to reduce hallucination~\cite{Chaudhari-2024-RLHF-arxiv} or further enhance the capacities of LLMs~\cite{rlmec,math-shepherd,wizardmath}.
Typically, a reward model will be trained on the preference data and leveraged to guide the reinforcement learning (RL) procedure~\cite{instructgpt,llama2,moss_rlhf}.
Proximal policy optimization (PPO) has been widely adopted in RLHF~\cite{ppo_a2c,moss_rlhf}.
Given the efficiency and expenses of the annotating process by human labeler, previous work has utilized the feedback from LLMs to instruct the RL process, named RLAIF~\cite{Bai-arXiv-2022-Constitutional,Yuan-arxiv-2024-Self}.
Furthermore, to prevent the instability of RL, a series of work~\cite{rdpo,orpo,simpo} utilized a similar objective function with SFT to model human preference.
Direct preference optimization (DPO)~\cite{dpo} is representative work of non-RL alignment.
In this work, we consider about how to unleash the potential of the non-RL method.

\paratitle{Unlearning of LLMs.}
Machine unlearning~\cite{Cao-ieee-2015-Towards,Bourtoule-arxiv-2019-MachineUnlearning,Wang-arxiv-2024-Machine,Chen-arxiv-2024-Machine} is an important technique for artificial intelligence systems to remove the knowledge about the restricted data (\eg unauthorized books), while keeping other knowledge and abilities of the systems.
To perform unlearning of LLMs, research has proposed several methods (\eg Gradient Ascent~\cite{Yao-arxiv-2023-Large,Maini-arxiv-2024-TOFU} and NPO~\cite{npo}), directly training LLMs on the invalid dataset to make LLMs forget relative knowledge.
Following the unlearning mechanism, in this work, we utilize an unlearning algorithm to correct the unaligned knowledge stored in the neurons of LLMs.

\paratitle{Low-Cost LLM Training.}
Due to the high computation cost of LLM training, previous work either trains a small portion of parameters in LLMs, or trains small-scale LLMs. 
For the former, existing work has shown the parameter redundancy in LLMs, and proposed related approaches to reduce the trainable parameters in LLM~\cite{lora,prefix-tuning}. 
Besides, quantization has also been used to further reduce the requirements of GPU memory~\cite{qlora}. 
Moreover, recent work has further improved the above methods by adjusting the updated parameters within LLMs or improving the training strategies~\cite{lora-drop,Du-arxiv-2024-unlocking}. 
For training small LLMs, to break the ceiling derived from the scaling law, knowledge distillation methods have been widely used for transferring the capabilities from the teacher model to small LLMs~\cite{teacherlm}. 
However, it is also promising to small models for hyper-parameters searching, to obtain the optimal training hyper-parameters for large LLMs~\cite{minicpm}. 
In this work, we mainly focus on training a small portion of parameters to reduce the useless redundancy in the human alignment training process. Such a way is capable of reducing the training cost and also improves the performance.

\section{Preliminary}

LLMs alignment refers to aligning the behaviors of LLMs to human preference, \eg helpfulness, honesty, and harmlessness~\cite{anthropic_3h}.
Existing work typically utilizes RLHF methods~\cite{Christiano-NeurIPS-2017-Deep} to fine-tune LLMs using human preference data, for improving alignment.
Formally, the human preference data is composed by input prompts, positive responses, and negative responses, denoted as $\mathcal{D}=\{\langle x_i, y_i^+,y_i^-\rangle\}_{i=1}^n$.
The input prompt or response consists of a series of natural language tokens $\{t_1,t_2,\dots,t_l\}$.
%Each element of data is natural language and can be considered as .
%In traditional alignment tasks, LLMs are 
Given the input prompt $x$, we aim to train LLMs that tend to generate the well-aligned positive response $y^+$, while avoiding generating the unaligned negative one $y^-$.
In this work, we focus on devising an effective training algorithm to improve the alignment of LLMs, which can be utilized to satisfy the diverse requirements in real world (\eg instruction following and question answering).
%By training on special datasets, LLMs can align with the human preference 
%Besides traditional alignment tasks, in this work, we also consider the complex reasoning tasks (\ie question-answering tasks and mathematical reasoning tasks), which can assess the helpfulness and honesty of LLMs.

According to our empirical study in Figure~\ref{intro}, updating only top-10\% trainable neurons would achieve better performance than full-parameter tuning for alignment training.
It indicates that there are redundant updates in the training process of LLMs, which may affect the alignment performance.
%we can observe that there are several redundant neurons in the neural networks, especially in LLMs.
%Inspired by the lottery ticket hypothesis~\cite{LotteryTicketHypothesis}, 
To address it, in this work, we aim to perform parameter-efficient fine-tuning for reducing the redundant updates on unrelated neurons, to improve the alignment of LLMs.
%in different downstream scenarios via only editing the most important neurons in LLM during alignment.
%Generally, existing alignment methods either utilize the training data to optimize the LLMs through gradient descent (\eg DPO~\cite{dpo}), or requires locating the relative neurons for each instance (\eg DINM~\cite{Wang-arxiv-2024-Detoxifying}).
%In the training procedure, former methods will update whole neurons in LLMs without pertinence, and later methods require frequently selecting neurons which might need huge expenses.
%In our work, we follow the existing alignment methods (\ie NPO~\cite{npo} and DPO~\cite{dpo}), and optimize the training pipeline to enhance training efficiency.
Given the training data, we first identify the highly-relevant neurons $\mathcal{N}=\{\theta_{i_1},\dots,\theta_{i_k}\}$ in the parameter matrices of LLMs, and perform low-redundant optimization on the LLM as:
\begin{equation}
% \small
\theta_i^{t+1} = \begin{cases} \text{Optimizer}(\theta_j^t, \nabla \theta_j^t),&\theta_j \in \mathcal{N} \\ \theta_j^{t},&\theta_j \notin \mathcal{N} \end{cases} ,
\end{equation}
where $\theta_j^t$ means the value of $j$-th neuron at the $t$-th step of training process, $\nabla \theta_j$ is the calculated gradient of $j$-th neuron for update. %through training loss $\mathcal{L}$.
%before the start of the training procedure.
%During the training process, data $\langle x_i, y_i^+,y_i^-\rangle$ is fed into LLMs, and the gradient of each neuron $\nabla \theta_j$ is calculated through training loss $\mathcal{L}$.
%Based on the gradient and important neuron set, the neurons in LLMs can be updated by the optimizer (\eg AdamW~\cite{adamw}) as follows, 

\begin{figure*}[t]
    \centering
    \includegraphics[width=0.98\textwidth]{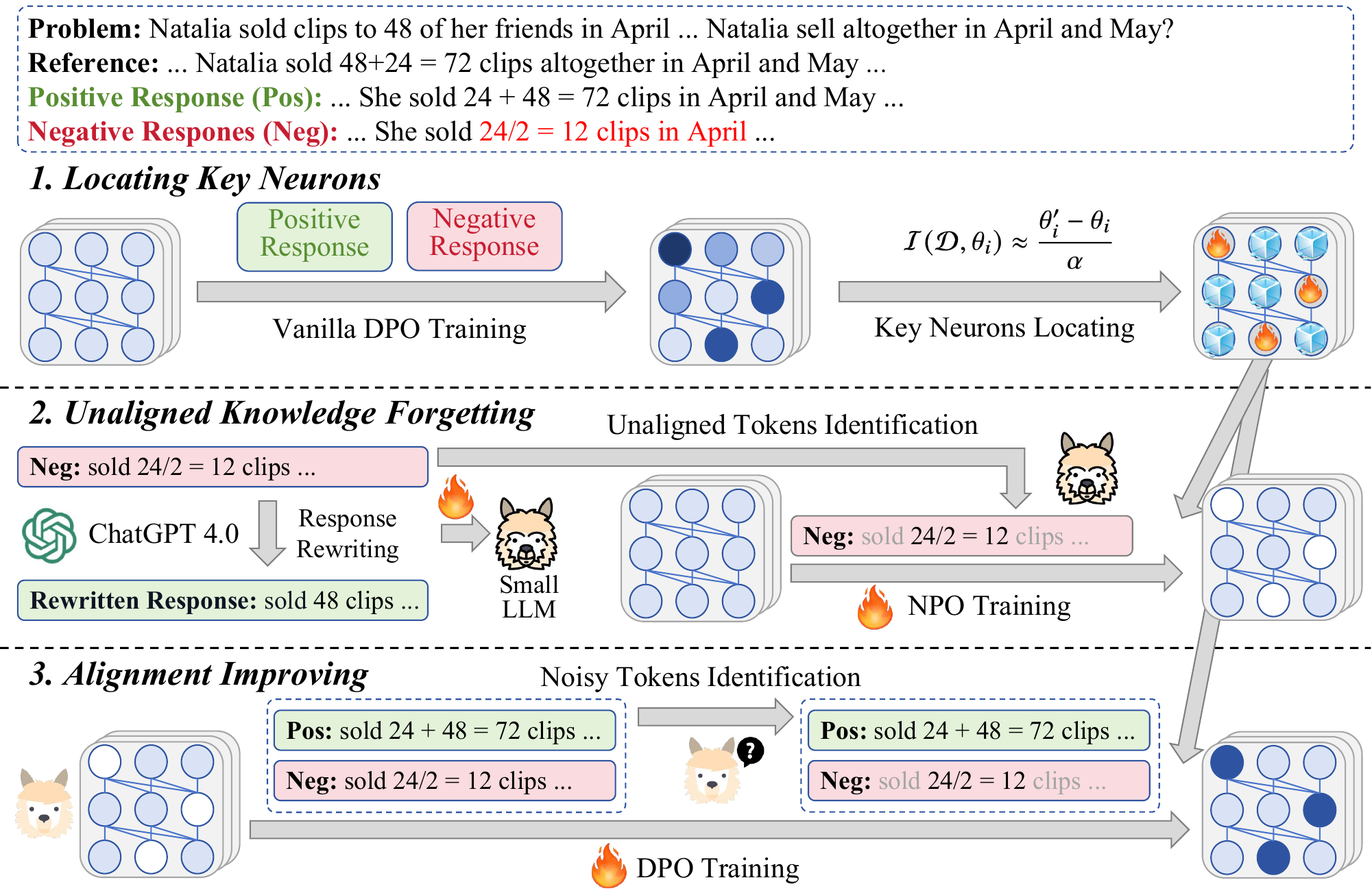}
    \caption{The framework of our proposed alignment method ALLO. We first locate the key neurons in LLMs by computing the weight changes of the reference model.% and selecting the top $k\%$ largest neurons. 
    Then, based on the selected key neurons, we perform a fine-grained unlearning using NPO to help LLMs forget unaligned knowledge, and fine-grained learning using DPO to further align LLMs to human preference.}
    \label{framework}
\end{figure*}

\section{Approach}
In this section, we introduce our proposed method ALLO, a low-redundant alignment method for fine-tuning LLMs.
In ALLO, we compute loss on selected key tokens, and only optimize the selected important neurons.
Concretely, we first train a reference model to locate the important neurons through gradient.
%fine-tune LLM and make it adapt to the downstream scenario, to 
Then, we identify the key tokens related to unaligned knowledge, and utilize the unlearning algorithm to update few neurons for forgetting them. 
%perform the unlearning process with fine-grained supervision signals on the original LLM to make it forget erroneous knowledge and unleash the potential of LLM.
Next, we leverage DPO algorithm to improve the alignment of the LLM, where the DPO reward is used for selecting the key tokens.
The framework of ALLO is presented in Figure~\ref{framework}.

\subsection{Locating Key Neurons}
%To reduce redundant update of training unnecessary parameters, 
We compute the importance of all the neurons for the human preference data to locate the related key neurons.
%Before starting the training process, we locate the key neurons, which store the knowledge and control the ability relative to downstream tasks.
We first train a reference model on the given data using DPO algorithm, and then design an efficient approximate estimation of the neuron importance based on its updated weights.

%To this end, we incorporate a gradient-based method to indicate the importance of each neuron.
%To perform this method, LLM is required to warm up via the scenario-relative data, and then we can obtain the gradient information for each neuron in LLM.

\paratitle{Training Reference Model.}
We train the reference model on the human preference data, to obtain the updated values of all neurons for importance estimation.
Thus, we select the same LLM as the backbone, and perform full-parameter fine-tuning using DPO algorithm on the entire dataset for one epoch.
The training objective is:
\begin{equation}
\small
    \mathcal{L}(d_i)=-\log\sigma\left(\beta\log\frac{P(y_i^+|x_i)}{P_{\text{ref}}(y_i^+|x_i)}-\beta\log\frac{P(y_i^-|x_i)}{P_{\text{ref}}(y_i^-|x_i)}\right), \label{eq-DPO}
\end{equation}
where $\beta$ is a hyper-parameter, and $d_{i} = \langle x_i, y_i^+, y_i^- \rangle$ is a training instance.
%To obtain the correlation of the downstream scenario and each neuron in LLM, we first warm up the LLM and then calculate the gradient information of each neuron.
%Given that the correctness of knowledge stored in LLM is difficult to identify, only utilizing the human-annotated reference to warm up LLM (\eg SFT) cannot locate the key neurons correctly.
For the scenarios that only one human feedback is provided, we regard it as the positive one, and leverage the response generated from LLM as the negative one.
%to construct the positive-negative instance pairs.
%Because DPO can synchronously optimize the correct knowledge and erroneous knowledge, LLM trained through DPO can update the value of neurons storing knowledge related to the corresponding task.

\paratitle{Neurons Importance Estimation.}
We aim to estimate the importance of each neuron for the given human preference dataset $\mathcal{D}$.
As LLMs are generally trained by gradient descent algorithm, the gradient value of a training instance $d_j$ on the neuron $\theta_i$ can reflect its influence on the neuron~\cite{Pruthi-2023-Estimating-neurips,Xia-2024-LESS-arxiv}, denoted as:
\begin{equation}
%\small
    \text{Influence}(d_j, \theta_i) \propto \nabla_{\theta_i}\mathcal{L}(d_j)
\end{equation}
%the gradient of the neurons in LLM during the training process can indicate the relationship between knowledge status in LLM and the knowledge required for specific tasks.
For human alignment, we use the DPO training loss in Eq.~\ref{eq-DPO} for influence estimation.
In this way, we can accumulate the gradients for all the instances from the human preference dataset, to estimate the influence of the dataset on the neuron.
Actually, the influence value also reflects the importance of the neuron for learning the dataset, as a large accumulated gradient value can denote more focus on training the neuron~\cite{Pruthi-2023-Estimating-neurips}.
As we adopt the gradient descent algorithm, the gradients for all the instances have been computed and subtracted in the one-epoch training process.
Thus, the difference between the neuron in the reference model $\theta'_i$ and original model $\theta_i$ can be regarded as the approximate value of the estimated importance score:
%Formally, for a dataset $\mathcal{D}=\{d_1,\dots,d_n\}$ and a training instance in the dataset $d_r=\langle x_r, y_r \rangle$, the L1 Norm of the gradient of LLM neurons can assess the influence of this instance on the corresponding neuron, denoted as follows,
%To a certain degree, given an instance in the dataset of the downstream scenario, we can measure the importance of each neuron in LLM through calculating the L1 Norm of the gradient in the neurons.
%However, one downstream task might require LLM to possess various knowledge and capacities.
%In this case, only utilizing one or few instances to produce the gradient and select key neurons cannot completely find out the relative neurons and might cause mistakes because of the bias of the instances.
%To address this issue, we leverage the accumulation of gradient in the training process to identify the importance of neurons, which can be regarded as the accumulation of influence of these training instances on the LLM neurons.
%Concretely, we select the instances from the corresponding scenario and make LLM fit to these instances.
%During the fitting process, various knowledge synchronously influences the neurons in LLM, and the accumulation of the gradient infers the importance of neurons.
%In summary, we leverage the following equation to calculate the importance of the neuron $\theta_i$ on the dataset $\mathcal{D}$, 
\begin{equation}
%\small
    \label{eq-locating}
    \mathcal{I}(\mathcal{D}, \theta_i) = \sum_{j=1}^{|\mathcal{D}|} \nabla_{\theta_i}\mathcal{L}(d_j) \approx \frac{\theta'_i - \theta_i}{\alpha},
\end{equation}
where %denote the weight of the neuron in the reference model and original model, respectively, 
$\alpha$ is the learning rate during DPO training.
Based on the estimated importance score, we can rank all the neurons and select the most important ones for training.
%value and gradient of $i-$th neuron at $t-th$ step in the training process respectively, and $T$ refers to the total training step.
%After the warming-up process, we obtain the accumulation of gradient of each neuron and the neurons which acquire top $k\%$ large importance will be selected into the key neurons set $\mathcal{N}$.
%Once the key neurons have been selected, during the following training process, only these selected neurons will be optimized.

\subsection{Unaligned Knowledge Forgetting}

For the forgetting stage, we utilize a token-level reward model that guides LLMs to focus on the tokens related to unaligned knowledge, and adopt a machine unlearning algorithm, \ie NPO~\cite{npo} that learns to forget them.

\paratitle{Unalignment-Related Tokens Identification.}
%Previous work has shown that there are several redundant tokens in the training content, which will affect the effectiveness of the training process~\cite{Lin-2024-Rho-arxiv}.
%Inspired by existing work~\cite{rlmec}, 
We train a token-level reward model to score tokens in the negative responses, according to their effect on unalignment.
Following existing work~\cite{rlmec}, we distill the capability of a strong LLM (\ie GPT-4~\cite{GPT-4}) to revise the unaligned response (to a well-aligned one) with minimum editing constraint, into a small LLM.
Then, we can utilize its output revision probability for each token, to compute the reward score as:
\begin{equation}
%\small
    \label{eq-identify_unaligned}
    r_{i,j} = \begin{cases} 1,&P_{re}(y_{i,j}|p_i, x_i, y_i^{+}, y^{-}_{i,<j})<u \\ 0,&\text{others} \end{cases},
    %~r'_{i,j} = P_{\text{GRM}}(\hat{s}_{i,j}|p_i,\tilde{s}_i,\hat{s}_i,\hat{s}_{i,<j}),
\end{equation}
%prompt the powerful LLM  to modify the generated response from the student LLM , and conduct the instance pairs to train a generative reward model (GRM).
where $p_i$ is the prompt to guide the reward model, $y_{i,j}$ is the $j$-th token in the negative response $y_i^{-}$, 
%generated response $\hat{s}_i$ from student LLM, the probability of GRM generating origin response $\hat{s}_i$ can be regarded as the correctness of current token.
%Formally, the reward of the $j-$th token in $i-$th training instance can be denoted as follows, 
$u$ is a hyper-parameter to control the threshold. 
%determine whether the current token is worthy to be learned by student LLM.
In this way, we can select the key tokens about the unalignment according to the 0-1 reward score.

\paratitle{Fine-grained Unlearning with NPO.}
Based on the selected unalignment-related key tokens, we perform unlearning to remove the unaligned knowledge in the LLM and unleash the potential of learning aligned knowledge.
Concretely, we utilize the NPO method, which is the revision based on DPO and only focuses on minimizing the likelihood of generating negative responses.
%and compares this likelihood of LLMs after and before training.
The objective function of NPO is as follows, 
\begin{equation}
% \small
    \mathcal{L}_{NPO}(\theta)={\log{\sigma\left(-\beta\log\frac{P(y_{i}^-|x_i)}{P_{\text{ref}}(y_i^-|x_i)}\right)}}.
\end{equation}
%as the backbone method for the unlearning stage.
Whereas, the NPO loss would also punish the tokens that are irrelevant to the unalignment but exist in the negative response.
%can not well handle the  is in instance level comparison between the generation probability of the student model and the reference model.
To address it, we constrain that only the key tokens are involved into loss computation, to avoid unlearning the irrelevant tokens.
%To provide fine-grained supervision, 
Formally, we decompose the objective into the token level, and add the 0-1 reward score as the token weights.
%the generation probability of generating the whole instance into generating each token, so that reward $r_{i,j}$ can be injected into the objective function.
Thus, the objective function can be revised as follows, and we only optimize the top-$k_1$\% most important neurons, denoted as $\mathcal{N}_1$,
\begin{equation}
\small
    \label{eq-train_npo}
    \mathcal{L}_{N}(\mathcal{N}_1)=-\sum_{j=1}^{l_i}{\log{\sigma\left(-\beta\log\frac{P(y_{i,j}^-|x_i,y_{i,<j}^-)}{P_{\text{ref}}(y_{i,j}^-|x_i,y_{i,<j}^-)}\times r_{i,j}\right)}}.
\end{equation}

\subsection{Alignment Improving}
For the learning stage, we further improve the alignment of the LLM that has unlearned the unaligned knowledge.
%In this part, we introduce the learning procedure of ALLO.
We adopt DPO~\cite{dpo} algorithm for training, and also leverage its computed reward score to distinguish the key tokens and noisy ones.

%Following the idea in the DPO algorithm, we leverage LLM itself as the reward model, to indicate whether the current token is redundant, which prevents LLM from learning useless information.

\paratitle{Noisy Tokens Identification.}
%In the DPO training process, the positive instance and negative instance will be optimized at the same time.
%However, existing work has proposed that most tokens in negative instances are credible and only a few tokens lead to the undesired result~\cite{}, showing that instance-level supervision signals might force LLM to learn the redundant tokens and information, leading to a decrease of LLM performance.
%In the learning stage, the LLM is trained to generate the positive responses while avoiding negative ones.
We also identify the noisy tokens in the negative responses using the reward score in DPO, for reducing their harmful influence on learning other key tokens.
As DPO requires to compare the token probabilities of the current-step LLM and its original probability, the reward of the key tokens initially own small values and increase smoothly.
However, the noisy ones typically lead to large reward values, and shock the training process~\cite{ppo_clip}.
%To address this problem, we eliminate potential redundant and noisy tokens via gradient masking.
%To identify these tokens, we follow the idea in existing work~\cite{dpo,Rafailov-2024-From-arxiv}, utilizing student LLM as the reward model.
Therefore, we can utilize the reward scores dynamically computed in the DPO process, to distinguish the key and noisy tokens, denoted as:
\begin{equation}
\small
    \label{eq-identify_noise}
    q_{i,j} = \begin{cases} 0,&r'_{i,j}\in\text{top } v\% \\ 1,&\text{others} \end{cases},~r'_{i,j}=\frac{P(y_{i,j}^-|x_i,y_{i,<j}^-)}{P_{\text{ref}}(y_{i,j}^-|x_i,y_{i,<j}^-)},
\end{equation}
where $v\%$ is the hyper-parameter to control the threshold. In this way, we can identify the noisy tokens causing abnormal large rewards with weight 0, and key tokens with weight 1.

%The ratio of the probability of generating negative instances between the student model before and after training can indicate the learning status of LLM on each token.
%The larger ratio refers to the confusion of the student model about the current token, which indicates that the corresponding token might contain noise and LLM should stop learning from this token.
%Therefore, we conduct the following equation to calculate the redundancy of each token and produce the gradient mask,
%where $x_i$, $y_i^+$, and $y_i^-$ denote the prompt, positive instance, and negative instance, respectively.
%In this equation, LLM would not learn from the redundant tokens, which might increase the training efficiency of the alignment process.

\paratitle{Fine-grained Learning with DPO.}
After obtaining the token weights, we also decompose the objective function of DPO into the token level, and add weights into the tokens from the negative response to provide fine-grained supervision.
%The gradient mask will only be added to the negative item in the DPO objective function.
%That is because optimizing the positive item and the negative item might cause redundancy, \ie synchronously increasing and decreasing the probability of generating similar content, which makes LLM confusing about the correctness of the corresponding knowledge.
%Through the gradient mask mechanism, LLM prevents learning from these noisy tokens, decreasing the redundancy in the training process.
Formally, the revised objective function is as follows:
\begin{equation}
\small
    \label{eq-train_dpo}
\begin{aligned}
    \mathcal{L}_{D}(\mathcal{N}_2)=-\log\sigma(&\beta\sum_{j=1}^{l_i^+}\log\frac{P(y_{i,j}^+|x_i,y_{i,<j}^+)}{P_{\text{ref}}(y_{i,j}^+|x_i,y_{i,<j}^+)}\\
    -&\beta\sum_{j=1}^{l_i^-}\log\frac{P(y_{i,j}^-|x_i,y_{i,<j}^-)}{P_{\text{ref}}(y_{i,j}^-|x_i,y_{i,<j}^-)}\times q_{i,j}),
\end{aligned}
\end{equation}
where we only optimize the top-$k_2$\% most important neurons, denoted as $\mathcal{N}_2$.
%In this function, LLM learns to generate the positive instance and decrease the probability of generating the undesired response without the influence of the redundant tokens.
\section{Experiment}

\subsection{Experimental Settings}

\begin{table}[t]
\small
    \centering
    \begin{tabular}{cccc}
        \toprule
        \textbf{Task} & \textbf{Train~/~Test} & \textbf{Dataset} & \textbf{Number} \\
        \midrule
        \multirow{3.5}*{IF} & Train & UltraFeedback & 23,976 \\
            \cmidrule{2-4}
            & \multirow{2}*{Test} & AlpacaEval 2.0 & 805 \\
                & & Arena-Hard & 500 \\
        \midrule
        \multirow{6.5}*{QA} & \multirow{2}*{Train} & ECQA & 7,598 \\
                & & QASC & 8,134 \\
            \cmidrule{2-4}
            & \multirow{4}*{Test} & ECQA & 2,194 \\
                & & QASC & 926 \\
                & & OBQA & 500 \\
                & & StrategyQA & 687 \\
        \midrule
        \multirow{5.5}*{Math} & Train & MetaMathQA & 40,000 \\
            \cmidrule{2-4}
            & \multirow{4}*{Test} & GSM8k & 1,319 \\
            &  & MATH & 5,000 \\
            &  & MAWPS & 2,065 \\
            &  & TabMWP & 1,000 \\
        \bottomrule
    \end{tabular}
    \caption{Statistics of the evaluation datasets. ``IF'' denotes the instruction following tasks.}
    \label{dataset}
\end{table}

In this section, we introduce the details of our evaluation process, including downstream datasets, baselines in the evaluation, and the implementation details of our proposed method.

\begin{table*}[ht]
    \small
    \centering
    \begin{tabular}{lcccccccccc}
        \toprule
        \multirow{2.5}*{\textbf{Methods}} & \multicolumn{5}{c}{\textbf{Question-Answering Tasks}} & \multicolumn{5}{c}{\textbf{Mathematical Reasoning Tasks}} \\
        \cmidrule(r){2-6}\cmidrule(r){7-11}
         & ECQA & QASC & OBQA & StrategyQA & Avg. & GSM8k & MATH & MAWPS & TabMWP & Avg.   \\
        \midrule
        SFT LLM & 69.92 & 55.51 & 52.60 & 55.75 & 58.45 & 55.9 & 11.8 & 79.9 & 56.7 & 51.1 \\
        \midrule
        + SFT & 69.14 & 55.40 & 49.80 & 59.24 & 58.40 & 56.2 & 11.8 & 80.0 & \underline{57.4} & 51.4 \\
        + RFT & 71.15 & 57.24 & 54.40 & 56.33 & 59.78 & 54.7 & 12.0 & 80.2 & 55.2 & 50.5 \\
        \midrule
        + DPO & 75.07 & 60.37 & 57.00 & 59.53 & 62.99 & 56.6 & 12.2 & 81.7 & 57.3 & 52.0 \\
        + R-DPO & \underline{75.52} & \underline{61.56} & \underline{58.40} & \underline{59.83} & \underline{63.83} & {56.9} & 12.3 & {82.3} & 57.2 & {52.2} \\
        + IPO & 47.86 & 43.20 & 41.80 & 43.38 & 44.06 & \textbf{58.0} & \underline{12.9} & \underline{82.4} & 55.5 & 52.2 \\
        + BCO & 68.87 & 55.18 & 45.40 & 57.21 & 56.67 & {57.2} & {12.4} & 81.8 & 56.3 & 51.9 \\
        + SimPO & 62.76 & 52.27 & 46.80 & 53.71 & 53.89 & \underline{57.9} & 12.8 & 82.1 & 56.7 & \underline{52.4} \\
        + NPO & 70.56 & 56.59 & 52.80 & 56.04 & 59.00 & 56.4 & 12.3 & 80.1 & 56.5 & 51.3 \\
        \midrule
        + Vanilla PPO & 70.65 & 55.29 & 53.40 & 56.33 & 58.92 & 55.2 & 11.6 & 79.4 & 56.5 & 50.7 \\
        + PPO A2C & 71.06 & 55.18 & 53.00 & 58.37 & 59.40 & 55.2 & 11.7 & 82.1 & 55.8 & 51.2 \\
        \midrule
        + ALLO  & \textbf{75.93} & \textbf{62.31} & \textbf{59.60} & \textbf{60.84} & \textbf{64.67} & 56.6 & \textbf{13.0} & \textbf{82.5} & \textbf{58.1} & \textbf{52.6} \\
        \bottomrule
    \end{tabular}
    \caption{Experimental results on question answering tasks and mathematical reasoning tasks. Avg. is the average accuracy of all sub-tasks. The best is denoted in bold and the second best is underlined.}
    \label{main_results_1}
\end{table*}

\begin{table}[ht]
    \small
    \centering
    \begin{tabular}{lccc}
        \toprule
        \multirow{2.5}*{\textbf{Methods}} & \multicolumn{3}{c}{\textbf{Instruction Following Tasks}} \\
        \cmidrule(r){2-4}
            & AlpacalEval 2.0 & Arena-Hard & Avg.  \\
        \midrule
        SFT LLM & 50.00 & 50.00 & 50.00 \\
        \midrule
        + SFT & 49.44 & 61.50 & 55.47 \\
        + RFT & 50.06 & 53.70 & 51.88 \\
        \midrule
        + DPO & 53.80 & 68.30 & 61.05 \\
        + R-DPO & 54.00 & \underline{72.20} & 63.10 \\
        + IPO & \underline{56.35} & 71.00 & \underline{63.68} \\
        + BCO & {54.79} & 71.80 & {63.30} \\
        + SimPO & 54.92 & 69.30 & 62.11 \\
        + NPO & 50.06 & 51.10 & 50.58 \\
        \midrule
        + Vanilla PPO & 48.75 & 48.20 & 48.48 \\
        + PPO A2C & 53.50 & 57.80 & 55.65 \\
        \midrule
        + ALLO  & \underline{55.78} & \textbf{74.90} & \textbf{65.34} \\
        \bottomrule
    \end{tabular}
    \caption{Experimental results on instruction following tasks. Avg. is the average win rate of all sub-tasks. The best are in bold and the second-best are underlined.}
    \label{main_results_2}
\end{table}

\paratitle{Datasets.}
We conduct the three downstream scenarios for the comprehensive evaluation, \ie question-answering (QA), mathematical reasoning, and instruction following. The statistics information of each task is presented in Table~\ref{dataset}.

$\bullet$ \textit{QA tasks} require LLMs to perform multi-step reasoning to solve problems. We adopt ECQA~\cite{ecqa}, QASC~\cite{qasc}, OpenbookQA~\cite{openbookqa}, and StrategyQA~\cite{strategyqa} as the evaluation tasks. LLMs are fine-tuned on the training set of ECQA and QASC to adapt to the QA tasks.

$\bullet$ \textit{Mathematica reasoning tasks} include four challenge tasks, \ie GSM8k~\cite{gsm8k}, MATH~\cite{math}, MAWPS~\cite{mawps}, and TabMWP~\cite{tabmwp}, containing problems with different levels of difficulty. To complete the mathematical knowledge and ability of LLMs, MetaMathQA~\cite{Yu-2023-MetaMath-arxiv} has been utilized to fine-tune the LLMs.

$\bullet$ \textit{Instruction following tasks} assess the capacity of LLMs to follow human instructions. AlpacaEval 2.0~\cite{alpacaeval} and Arena-Hard~\cite{arenahard} are considered as the downstream tasks. We adopt the alpaca dataset~\cite{alpaca} to fine-tune the base LLMs and UltraFeedback dataset~\cite{ultrafeedback} for the further training process (\eg, DPO, ALLO).

For QA tasks and mathematical tasks, accuracy has been adopted as the evaluation metric.
For the instruction following tasks, we employ \texttt{gpt-3.5-turbo} as the judge model and report the win rate over the backbone model (\ie SFT LLM).

\paratitle{Baselines.}
We incorporate three categories of methods in the evaluation, including supervised fine-tuning (\ie SFT~\cite{instructgpt} and RFT~\cite{rft}), reinforcement learning (\ie Vanilla PPO~\cite{ppo} and PPO A2C~\cite{ppo_a2c}), and alignment without RL (\ie DPO~\cite{dpo}, R-DPO~\cite{rdpo}, IPO~\cite{ipo}, BCO~\cite{bco}, SimPO~\cite{simpo}, and NPO~\cite{npo}).

\paratitle{Implementation Details.}
In the experiment, we fine-tune LLaMA 2 7B~\cite{llama2} on instruction datasets corresponding to the downstream scenarios to obtain the backbone model (\ie SFT LLM), and conduct further training processes based on this model in the evaluation.
The details of hyper-parameters are presented in Table~\ref{hyper_parameters}.

\begin{table*}[t]
\small
    \centering
    \begin{tabular}{cccccccccc}
        \toprule
        \multicolumn{2}{c}{\textbf{Forgetting Stage}} & \multicolumn{2}{c}{\textbf{Learning Stage}} & \textbf{QASC} & \textbf{OBQA} & \textbf{MATH} & \textbf{MAWPS} & \textbf{AlpaceEval 2.0} & \textbf{Arena-Hard}  \\
        \cmidrule(r){1-2}\cmidrule(r){3-4}\cmidrule(r){5-5}\cmidrule(r){6-6}\cmidrule(r){7-7}\cmidrule(r){8-8}\cmidrule(r){9-9}\cmidrule(r){10-10}
        TLR & Mask & TLR & Mask & Acc. (\%) & Acc. (\%) & Acc. (\%) & Acc. (\%) & WR (\%) & WR (\%) \\
        \midrule
        \ding{52} & Top-k & \ding{52} & Top-k & 62.31 & 59.60 & 13.0 & 82.5 & 55.78 & 74.90 \\
        \midrule
        \ding{52} & Top-k & \ding{55} & Top-k & 62.42 & 59.00 & 12.4 & 82.5 & 55.60 & 75.10 \\
        \ding{55} & Top-k & \ding{52} & Top-k & 61.56 & 58.20 & 12.7 & 82.3 & 55.47 & 73.20 \\
        \midrule
        \ding{52} & \ding{55} & \ding{52} & Top-k & 61.77 & 58.80 & 13.1 & 82.4 & 55.22 & 73.20 \\
        \ding{52} & Top-k & \ding{52} & \ding{55} & 61.66 & 58.20 & 12.7 & 81.7 & 53.98 & 69.70 \\
        \ding{52} & Last-k & \ding{52} & Top-k & 62.20 & 59.40 & 12.5 & 82.5 & 55.29 & 70.80 \\
        \ding{52} & Top-k & \ding{52} & Last-k & 61.77 & 59.00 & 10.2 & 70.9 & 51.74 & 61.20 \\
        \midrule
        - & - & \ding{52} & Top-k & 62.20 & 59.20 & 12.3 & 82.2 & 55.60 & 72.60 \\
        \ding{52} & Top-k & - & - & 56.16 & 53.20 & 11.8 & 79.9 & 51.37 & 50.20 \\
        \bottomrule
    \end{tabular}
    \caption{The results of ablation study. ``Acc.'' and ``WR'' denote accuracy and win rate, respectively. ``TLR'' denotes the whether adopting token-level rewards in each stage. ``Mask'' indicates the neuron masking mechanism.}
    \label{ablation}
\end{table*}

\subsection{Main Results}
The results of ALLO and baseline approaches in our evaluation are presented in Table~\ref{main_results_1} and Table~\ref{main_results_2}.

According to the evaluation, we can observe that ALLO outperforms other baselines in almost all downstream scenarios and makes a great improvement over NPO and DPO, which are the backbone methods of ALLO.
That is because ALLO makes great efforts to reduce the redundant elements in the alignment process, including neurons in LLMs and tokens in training data.
Experimental results have shown the effectiveness of ALLO.

Besides, comparing the performance between the algorithm with fine-grained supervision signals (\eg ALLO, PPO A2C) and the algorithm without them (\eg DPO, Vanilla PPO), the effectiveness of the fine-grained supervision signals has been verified.
Specifically, PPO A2C has achieved a 55.65\% average win rate in instruction following tasks, while Vanilla PPO only achieved 48.48\%.
Instance-level supervision cannot focus on the details in the training data, which will optimize the erroneous parts and hurt the performance of the training methods.
In contrast, token-level supervision signals can better identify whether the token is worthy to be learned, which reduces the redundancy of training content.

Moreover, the improvement brought by the unlearning method (\ie NPO) has demonstrated that aligned and unaligned knowledge are both stored in LLMs.
In the training process of NPO, the LLMs are not exposed to new knowledge and new capacities, and only are guided to forget the unaligned knowledge.
This phenomenon further verifies the importance of the unlearning stage and the existence of redundant neurons in LLMs.
Without the redundant neurons, is difficult of LLMs to learn both aligned and unaligned knowledge simultaneously.

Finally, we can observe that ALLO outperforms DPO and its various (\eg R-DPO, SimPO) in all downstream scenarios, especially in the instruction following tasks. 
This is because DPO and its various guide LLMs to learn the positive and negative instances simultaneously, which will make LLMs confused about the aligned components in the negative instances.
In contrast, ALLO first utilizes the unlearning process to lose the probability distribution in LLMs and leverage the fine-grained supervision signals to indicate the redundant tokens in the training data, to enhance the training efficiency.

\begin{figure}[t]
    \centering
    \includegraphics[width=0.48\textwidth]{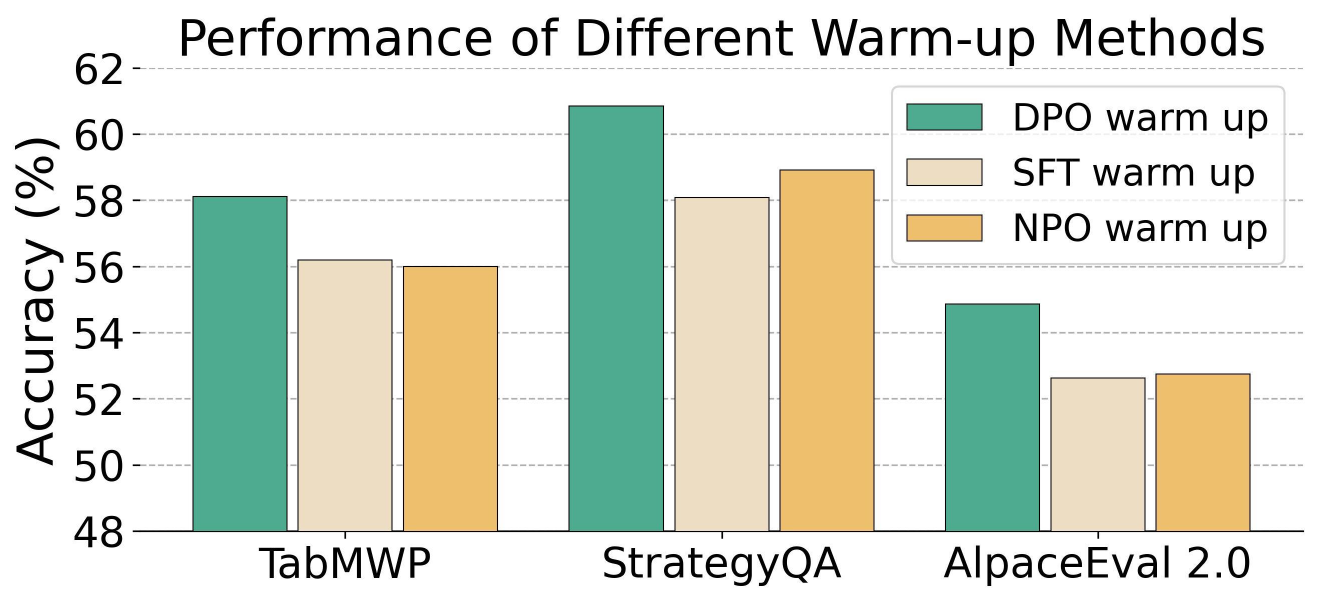}
    \caption{The experimental results of the influence of different warm-up methods on downstream tasks.}
    \label{warmup_method}
\end{figure}

\subsection{Detailed Analysis}

To further analyze our proposed ALLO, we conduct the ablation study, and analyze the influence of different warm-up methods and neuron mask ratio.
Besides, we present a case study in Appendix~\ref{case_study}.

\paratitle{Ablation Study.}
To assess the effectiveness of each module in ALLO, we conduct the ablation study and present the evaluation results in Table~\ref{ablation}.
According to the results, we can observe that removing any component of ALLO will hurt the performance, which has verified each module in ALLO is necessary and contributes to the final results of ALLO.
Besides, in QA tasks, the results of removing the neuron mask and adopting the Last-k neuron mask indicate the existence of redundant neurons in LLMs, which is the same as our empirical study.
For details, even adopting the Last-k neuron mask in Stage 2 (\eg 59.00\% accuracy of OBQA) outperforms the variant without neuron masking (\eg 58.20\% accuracy of OBQA).
That is because training the whole neurons in LLMs will decrease the training efficiency, and redundant updates affect the performance of downstream tasks.
Moreover, without the forgetting stage, ALLO still performs better than DPO in most tasks.
The reason is that the token-level reward and the neuron masking mechanism reduce the redundancy and make the training process focus on effective details in the training instances, making better utilization of the information in the dataset.

\begin{figure}[t]
    \centering
    \includegraphics[width=0.48\textwidth]{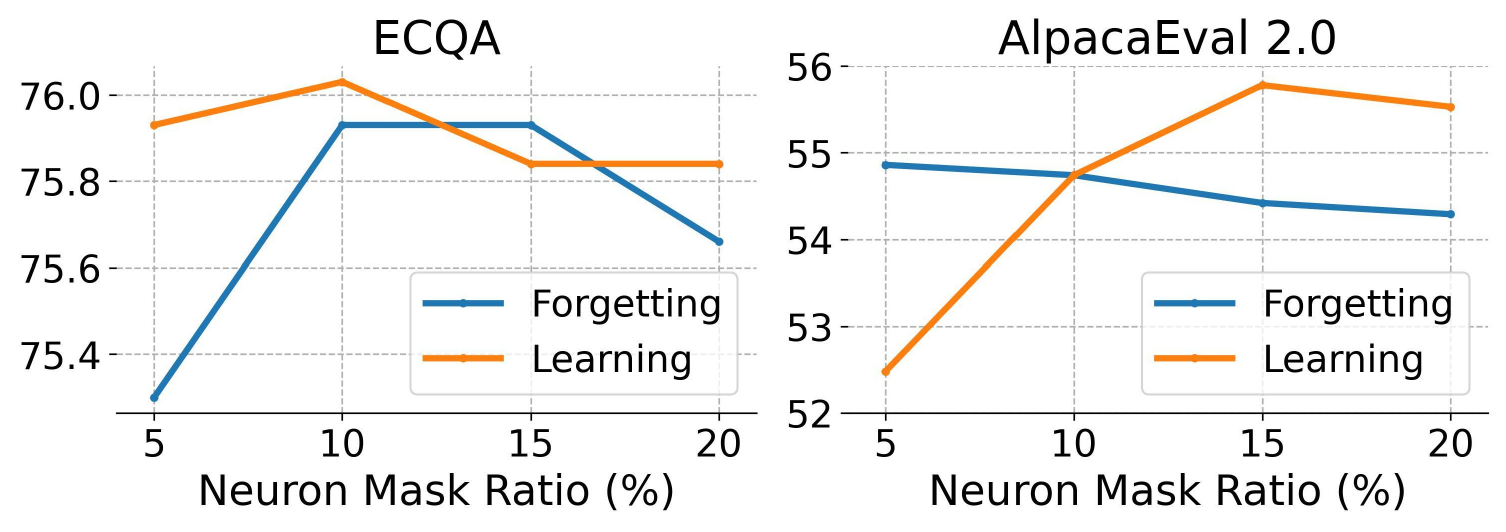}
    \caption{The experimental results of the different neuron mask ratios on ECQA and AlpaceEval 2.0, reporting the accuracy and win rate respectively. In the evaluation, we keep the mask ratio of one stage frozen and change the ratio of another stage.}
    \label{mask_ratio}
\end{figure}

\paratitle{Influence of Different Warm-up Methods.}
To assess the influence of different warm-up methods (\ie DPO, SFT, and NPO), we conduct the relative experiment and present the results in Figure~\ref{warmup_method}.
In all of the evaluation tasks, leveraging DPO to warm up LLMs and select important neurons has achieved the best performance that other warm-up methods.
Whether SFT or NPO, these training methods only utilize a single part of the training dataset, \ie the positive responses or the negative responses, respectively.
However, positive responses indicate the knowledge that LLMs should possess, and negative responses can locate unaligned knowledge stored in LLMs.
These responses are both important and necessary in selecting the key neurons for the corresponding scenario.
DPO can leverage the information in this data and guide LLMs to learn the aligned knowledge and eliminate unaligned one.
In this warm-up process, the neurons related to downstream tasks will be modified largely, causing the large value of the gradient, which can more precisely locate the important neurons for the following training process.

\paratitle{Analysis the Ratio of Neuron Mask.}
We present the results of different ratios of neuron masks on the QA task (\ie ECQA) and the instruction following task (\ie AlpaceEval 2.0) in Figure~\ref{mask_ratio}.
According to the evaluation results, we can observe that the performance first increases and then decreases, with the change of the neuron mask ratio.
Concretely, for the ECQA task, selecting 10\% neurons in the learning stage achieves the best performance, while selecting fewer or more neurons will hurt the accuracy of LLMs on downstream tasks.
The increasing stage indicates that there are still several important neurons not been selected, which affects LLMs learning task-specific knowledge and abilities.
After the increasing stage, the selected neurons set $\mathcal{N}$ contains more and more redundant neurons, interfering the learning process of other neurons and hurting the performance of the LLMs.
The evaluation results have verified the existence of redundant updates in LLM alignment and shown that training an appropriate amount of neurons can reduce the redundancy and enhance the performance of LLMs.
\section{Conclusion}

In this paper, we proposed ALLO, an alignment method with low-redundant optimization, to train the most related neurons with the most useful supervised signals.
%leveraging low-redundant optimization to eliminate the redundant updates in the LLM alignment process.
In ALLO, we first estimated the importance of neurons in the LLM based on the weight changes of a reference model, and located the most related neurons for optimization.
%utilized the accumulated value of the gradient to estimate the importance of neurons in LLMs and selected the most relative neurons of downstream scenarios, only which will be optimized in the following training process.
Then, we decomposed the alignment process into the forgetting and learning stages, where we leveraged token-level reward and DPO reward scores to identify the key tokens, and computing loss on them for training.
%produce the fine-grained supervision signals in the training process, focusing on reducing redundant tokens in the training data.
Experimental results on question-answering tasks, mathematical reasoning tasks, and instruction following tasks have shown the effectiveness of ALLO.

As future work, we will consider leveraging ALLO on other important scenarios, \eg reducing hallucination. Besides, we will also implement ALLO in larger LLMs and multimodal LLMs to validate its effectiveness.

\section*{Limitations}

In this section, we discuss the limitations of our work.
First, we only conduct the experiment of ALLO on 7B LLMs, with the evaluation of the LLMs with larger scaling of parameters, because of the limitation of computation resources.
Actually, we comprehensively assess the performance of ALLO and the existing competitive baseline methods in various downstream tasks, and the experiment results have verified the effectiveness of our proposed methods.
Second, we adopt complex reasoning and human alignment tasks in our evaluation, which mainly assess the helpfulness of LLMs.
The performance of ALLO on other aspects, \eg reducing hallucination and generating harmless response, has not been verified in this work.
We leave it as future work.
Finally, we do not consider the potential risk of ethics risk during LLM deployment and will investigate this issue in the future.

\section*{Acknowledgement}
This work was partially supported by National Natural Science Foundation of China under Grant No. 62222215, Beijing Natural Science Foundation under Grant No. L233008 and 4222027. Xin Zhao is the corresponding author.

% \section*{Ethics Statement}
% Scientific work published at EMNLP 2023 must comply with the \href{https://www.aclweb.org/portal/content/acl-code-ethics}{ACL Ethics Policy}. We encourage all authors to include an explicit ethics statement on the broader impact of the work, or other ethical considerations after the conclusion but before the references. The ethics statement will not count toward the page limit (8 pages for long, 4 pages for short papers).

\bibliography{anthology,custom}
\bibliographystyle{acl_natbib}

\newpage

\appendix

\begin{algorithm}[t]
\small
\caption{The ALLO algorithm.}
\label{code_allo}
\SetKwInOut{Input}{Input}
\SetKwInOut{Output}{Output}

\Input{Training set $\mathcal{D}=\{\langle x_i,y_i^+,y_i^- \rangle\}_{i=1}^{n}$, the teacher model (GPT-4o), and the SFT model $\theta_{\text{SFT}}$.}
\Output{A well-aligned model $\theta$.}

\BlankLine
\tcp{1. Locating Key Neurons}
$\theta' \leftarrow DPO(\theta_{SFT})$\;
\For{each neuron $\theta_i$ in warmed up model $\theta'$}{
    Calculate the importance of $\theta_i$ using Eq.~\ref{eq-locating}\;
}
Sort the importance of each neuron\;
Select the top-k relative neurons into $\mathcal{N}$\;

\BlankLine
\tcp{2. Unaligned Knowledge Forgetting}
\For{each instance $\langle x_i,y_i^+,y_i^- \rangle$ in $\mathcal{D}$}{
    \If{the data is sampled}{
        The teacher model rewrites the negative response $y_i^-$\;
    }
}
Leverage the rewritten response to fine-tune the small LLM to obtain the $\theta_{rm}$\;
\For{each instance $\langle x_i,y_i^+,y_i^- \rangle$ in $\mathcal{D}$}{
    Identify the unaligned token using Eq.~\ref{eq-identify_unaligned}\;
    Optimize the neurons in $\mathcal{N}$ using Eq.~\ref{eq-train_npo}\;
}
Obtain the model $\theta_{\text{forget}}$ forgetting unaligned knowledge\;

\BlankLine
\tcp{3. Alignment Improving}
\For{each instance $\langle x_i,y_i^+,y_i^- \rangle$ in $\mathcal{D}$}{
    Identify the noise token using Eq.~\ref{eq-identify_noise}\;
    Optimize the neurons in $\mathcal{N}$ using Eq.~\ref{eq-train_dpo}\;
}
Obtained the well-aligned model $\theta$\;
\end{algorithm}

\begin{table*}[ht]
    \small
    \centering
    \begin{tabular}{ccccc}
        \toprule
        Stage & Hyper-Parameter & Question-Answering & Mathematical Reasoning & Human Alignment \\
        \midrule
        \multirow{4}*{Stage 1} & Learning Rate & $1\times 10^{-7}$ & $5\times 10^{-8}$ & $1\times 10^{-7}$ \\
        & Batch Size & 32 & 512 & 128 \\
        & Selected Neuron Ratio & 5\% & 5\% & 10\% \\
        & Threshold $u$ & 0.95 & 0.95 & 0.95 \\
        & $\beta$ in NPO & 0.1 & 0.1 & 0.1 \\
        \midrule
        \multirow{5}*{Stage 2} & Learning Rate & $5\times 10^{-6}$ & $1\times 10^{-6}$ & $5\times 10^{-6}$ \\
        & Batch Size & 32 & 512 & 128 \\
        & Selected Neuron Ratio & 10\% & 20\% & 15\% \\
        & Threshold $v$ & 20\% & 50\% & 20\% \\
        & $\beta$ in DPO & 0.1 & 0.1 & 0.1 \\
        \bottomrule
    \end{tabular}
    \caption{The details of hyper-parameters in the evaluation.}
    \label{hyper_parameters}
\end{table*}

\begin{table*}[t]
    \small
    \centering
    \begin{tabular}{ll}
        \toprule
        \multicolumn{1}{l}{\begin{tabularx}{0.1\textwidth}{@{}X@{}}
            \textbf{Distillation for Solution Rewriting}
        \end{tabularx}} & \multicolumn{1}{l}{\begin{tabularx}{0.8\textwidth}{@{}X@{}}
            Given the problem and the correct solution, you need to correct the mistakes in the prediction to get the correct answer. You should write down the correct prediction and use ``The answer is: '' (without quotation mark) to identify the final answer. You should make minimal modifications. You should not copy the problem.\\ \\ \#\#\# Problem:\\ \{\texttt{Problem}\} \\ \\ \#\#\# Correct solution:\\ \{\texttt{Ground-Truth Reference}\}\\ \\ \#\#\# Prediction:\\ \{\texttt{Generated Response from LLM}\} \\ \\ \#\#\# Correct prediction:\\
        \end{tabularx}} \\
        \midrule
        \multicolumn{1}{l}{\begin{tabularx}{0.1\textwidth}{@{}X@{}}
            \textbf{Prompt for Downstream Tasks}
        \end{tabularx}} & \multicolumn{1}{l}{\begin{tabularx}{0.8\textwidth}{@{}X@{}}
            Below is an instruction that describes a task. Write a response that appropriately completes the request.\\ \\ \#\#\# Instruction:\\ \{\texttt{Problem to be solved}\} \\ \\ \#\#\# Response: Let's think step by step.
        \end{tabularx}} \\
        \bottomrule
    \end{tabular}
    \caption{The template of prompt in ALLO.}
    \label{prompt}
\end{table*}

\begin{table*}[ht]
    \small
    \centering
    \begin{tabular}{ll}
        \toprule
        \multicolumn{1}{l}{\begin{tabularx}{0.1\textwidth}{@{}X@{}}
            \textbf{Problem}
        \end{tabularx}} & \multicolumn{1}{l}{\begin{tabularx}{0.8\textwidth}{@{}X@{}}
            John needed to go across the country so he ordered tickets online and got into his car.  Where might he be going?\\Options:\\train station\\opera\\car rental shop\\show\\race track
        \end{tabularx}} \\
        \midrule
        \multicolumn{1}{l}{\begin{tabularx}{0.1\textwidth}{@{}X@{}}
            \textbf{Ground-Truth Solution}
        \end{tabularx}} & \multicolumn{1}{l}{\begin{tabularx}{0.8\textwidth}{@{}X@{}}
            He was going to train station so he got into his car. \\ John needed to go across the country by train so he ordered tickets online. \\ The answer is train station
        \end{tabularx}} \\
        \midrule
        \multicolumn{1}{l}{\begin{tabularx}{0.1\textwidth}{@{}X@{}}
            \textbf{Generated Solution Before Training}
        \end{tabularx}} & \multicolumn{1}{l}{\begin{tabularx}{0.8\textwidth}{@{}X@{}}
            Race tracks are in the outskirts of the city \\ One can go to a race track by car \\ \textcolor{red}{Race tracks are in the country side.} \\ The answer is race track
        \end{tabularx}} \\
        \midrule
        \multicolumn{1}{l}{\begin{tabularx}{0.1\textwidth}{@{}X@{}}
            \textbf{Generated Solution After DPO}
        \end{tabularx}} & \multicolumn{1}{l}{\begin{tabularx}{0.8\textwidth}{@{}X@{}}
            A person may go to another place by car \\ When someone orders ticket online, it means that he is going to travel \\ When someone goes from one place to another by car, he crosses the country \\ When someone crosses the country, he might be \textcolor{red}{going to the other end of the country}. \\ The answer is race track
        \end{tabularx}} \\
        \midrule
        \multicolumn{1}{l}{\begin{tabularx}{0.1\textwidth}{@{}X@{}}
            \textbf{Generated Solution After ALLO}
        \end{tabularx}} & \multicolumn{1}{l}{\begin{tabularx}{0.8\textwidth}{@{}X@{}}
            John might be \textcolor{red}{going to the train station as he ordered tickets online} and got into his car. \\ John needs to go across the country so he might be going to the train station. \\ The answer is train station
        \end{tabularx}} \\
        \bottomrule
    \end{tabular}
    \caption{The case study for question-answering tasks.}
    \label{case_study_qa}
\end{table*}

\section{Algorithm of ALLO}

We present the pipeline of ALLO in Algorithm~\ref{code_allo}.
The process of ALLO includes three stages, \ie locating key neurons, unaligned knowledge forgetting, and alignment improving.

\section{Details of Hyper-Parameters}

To better understand and reproduce our proposed ALLO, we presented the hyper-parameters in ALLO in Table~\ref{hyper_parameters}.
The hyper-parameters are a little different between different downstream tasks, that is because these tasks are in different difficulty levels and require different abilities of LLMs.
It should be noted that, to conduct a fair comparison, the hyper-parameters of baseline methods are also adjusted to adapt to the corresponding tasks for better performance.

\section{Prompt Templates of ALLO}

In ALLO, we utilize prompts to guide the teacher model to rewrite the generated response from student models and induce the student model to solve the downstream tasks.
The templates of the prompt in ALLO are presented in Table~\ref{prompt}.
For the solution rewriting process, we feed the problem, ground-truth reference, and generated response into the teacher model, with the instruction of rewriting in the prefix.
Besides, for the downstream tasks, the instruction prefix and problem will be given into LLMs.

\section{Case Study}
\label{case_study}

To better demonstrate our proposed ALLO, we present the case study on QA task (\ie ECQA) in Table~\ref{case_study_qa}.
In this case, we can observe that LLM after DPO training still cannot catch the relation between ``tickets'' and the destination John needed to go to, and focus on the relation between ``cross country'' and ``race track''.
This phenomenon has shown that unaligned knowledge is not eliminated and still exists in LLMs after DPO training.
In contrast, after ALLO training, LLM can correctly seize on the key elements of the problem (\ie ``ticket'') and perform reasoning along the correct direction.
That is because low-redundant optimization can reduce the redundant updates in the alignment process and make LLMs focus on the key knowledge and information.

\end{document}